\documentclass[journal]{IEEEtran}
\usepackage[utf8]{inputenc} 
\usepackage[T1]{fontenc}    
\usepackage{hyperref}       
\usepackage{url}            
\usepackage{booktabs}       
\usepackage{amsfonts}       
\usepackage{nicefrac}       
\usepackage{microtype}      

\usepackage{amsmath,bm}

\usepackage{mathtools}
\usepackage{subcaption}
\usepackage{amssymb}
\usepackage[shortlabels]{enumitem}
\usepackage{graphicx}
\usepackage{color}
\usepackage{schemata}
\usepackage[ruled,vlined]{algorithm2e}
\makeatletter
\newcommand{\algorithmstyle}[1]{\renewcommand{\algocf@style}{#1}}
\makeatother
\usepackage{adjustbox}
\usepackage{float}
\usepackage{amsthm}
\usepackage{chngcntr}
\usepackage{apptools}
\usepackage{thmtools}
\usepackage{thm-restate}
\usepackage{cleveref}
\usepackage{wrapfig}
\usepackage{enumitem}

\newtheorem{assumption}{Assumption}

\newcommand{\zred}{\textcolor{black}}

\begin{document}

\title{Towards Optimal Pricing of Demand Response - A Nonparametric Constrained Policy Optimization Approach}

\author{{Jun Song,
        Chaoyue Zhao,~\IEEEmembership{Member,~IEEE}}

\thanks{Jun Song and Chaoyue Zhao are with University of Washington, E-mail:juns113@uw.edu; cyzhao@uw.edu. This work is funded by NSF ECCS-2046243}
}


\maketitle

\begin{abstract}
Demand response (DR) has been demonstrated to be an effective method for reducing peak load and mitigating uncertainties on both the supply and demand sides of the electricity market. One critical question for DR research is how to appropriately adjust electricity prices in order to shift electrical load from peak to off-peak hours. In recent years, reinforcement learning (RL) has been used to address the price-based DR problem because it is a model-free technique that does not necessitate the identification of models for end-use customers. However, the majority of RL methods cannot guarantee the stability and optimality of the learned pricing policy, which is undesirable in safety-critical power systems and may result in high customer bills. In this paper, we propose an innovative nonparametric constrained policy optimization approach that improves 
optimality while ensuring stability of the policy update, by removing the restrictive assumption on policy representation that the majority of the RL literature adopts: the policy must be parameterized or fall into a certain distribution class. We derive a closed-form expression of optimal policy update for each iteration and develop an efficient on-policy actor-critic algorithm to address the proposed constrained policy optimization problem. The experiments on two DR cases show the superior performance of our proposed nonparametric constrained policy optimization method compared with state-of-the-art RL algorithms. 
\end{abstract}
\IEEEpeerreviewmaketitle

\section{Introduction}
Demand response (DR), as an efficient way to reduce the peak load and improve the grid reliability, has been widely applied in recently years and recognized as one of the main forces to advance the smart grid technology. In order to incorporate DR resources into the wholesale energy market, a number of regional grid operators (such as NYISO, PJM, ISO-NE, and ERCOT) have established demand response programs for consumers to participate \cite{2021assessment}. Federal Energy Regulatory Commission (FERC) also issued its order in 2020 to enable distributed energy resources, which include demand response, to participate in regional wholesale electricity markets \cite{2020order}.

In general, incentive-based programs and price-based programs are the two categories of DR systems. In incentive-based schemes, customers receive fixed or time-varying compensation in exchange for limiting their power usage during peak hours or system contingencies. In price-based schemes, customers are encouraged to shift their electricity consumption from high to low price periods in order to save money on their electricity bill. In the latter case, deriving an effective and efficient pricing strategy is critical in order to adjust electricity consumption in a reliable and timely manner.

Traditionally, a Stackelberg game model \cite{maharjan2013dependable,yu2015real} is often used to characterize the interactions between DR service provider (SP) and participating consumers (PCs), in which SP acts as the leader who sets the price strategy and PCs are the followers who adjust their electricity consumption. However, in practice, it is very challenging for both SP and PCs to set up each other’s model and accurately estimate the model parameters. Also, due to a possible large number of PCs, there will be a significant number of models with different customer behavior patterns, which makes the model learning even harder. This challenge can be addressed by reinforcement learning (RL), since it is a model-free technique that does not require the identification of models for individual players.





In this vein, a variety of model-free RL methods including value-based methods such as Q-learning \cite{kim2015dynamic,remani2018residential,lu2018dynamic} and policy-based methods such as Advantage Actor Critic (A2C) \cite{bahrami2017online}, Trust Region Policy Optimization (TRPO) \cite{li2020real} have been exploited to address the dynamic pricing strategy of DR problems. However, the performances of traditional model-free RL methods are not satisfactory in practice for the following reasons: 1) different behavior patterns of PCs bring a high degree of environmental dynamics, which inevitably induces the instability of policy update and the difficulty for the convergence of final policy; 2) many RL methods may not reach policy optimality even after it stabilizes. In addition to bad initialization and inadequate
exploration, the model restriction that is employed could exclude the algorithms from attaining a more optimal policy \cite{tessler2019distributional}. 

In this paper, we propose a novel trust region constrained policy optimization method that can effectively address the DR pricing strategy problem. Similar as TRPO, we  impose a Kullback-Leibler
(KL) divergence based constraint (trust region constraint) to restrict the size of the policy update, so that a good stability can be maintained. However, unlike the traditional gradient based policy optimization methods such as A2C, TRPO, or PPO \cite{schulman2017proximal} that limit the policy
representation to a particular parametric distribution class (e.g., Gaussian \cite{schulman2015trust}), we release this restrictive distributional assumption by allowing all admissible policies in the trust region. Since in this case, the policy learnt is not confined to the scope of parametric functions, this
certainly opens up the possibility of converging to a better final policy. In fact, we observed significant improvements in nearly all our test cases. In addition, the proposed method is more flexible than the majority of the state-of-the-art gradient based policy optimization methods - it can be applied to both discrete and continuous action cases. We successfully derive the reformulations of the proposed policy optimization problem and obtain the closed-form optimal policy update. We have investigated the effectiveness of the proposed approach with two representative test cases on DR. Both cases demonstrate that our approach outperforms state-of-the-art gradient based policy optimization methods.


\section{MDP Formulation}
We use a similar Markov Decision Process (MDP) model as \cite{lu2018dynamic} to describe the dynamic decision-making problem of price-based DR program, which is defined as a tuple $<T, \lambda, \mathcal{S}, \mathcal{A}, \bm{p}, \bm{r}>$, where $T$ is the time horizon, $\lambda$ is the discount factor, $\mathcal{S}$ is the state space, $\mathcal{A}$ is the action space, $\bm{p}$ is the transition probability between states depending on the taken action, and $\bm{r}$ is the reward associated with the taken action. However, unlike \cite{lu2018dynamic}, we allow continuous state and action for our MDP model.

\begin{itemize}
\item {\bf Time Horizon}: We consider a finite time horizon with length $T = 24$, which represents $24$ hours in a day. We use $t = 1, 2, \dots, T$ to represent the discrete time stage. 

\item {\bf Action and State}: 
The action of the dynamic pricing DR model is defined as $a_t = (\phi_{t,1}, \dots, \phi_{t,N})$, where $N$ is the total number of PCs and $\phi_{t,n}$ denotes the retail electricity price for the $n$-th PC at time $t$. 

The state of the dynamic pricing DR model is defined as 
\begin{equation*}
s_t = ((d_{t,1}, c_{t,1}), \dots, (d_{t,N}, c_{t,N})), \end{equation*}
where $d_{t,n}$ represents the base energy demand of the $n$-th PC at time $t$ before knowing the retail price, and $c_{t,n}$ represents the actual energy consumption of the $n$-th PC at time $t$ after knowing the retail price. 

The base energy demand and the actual energy consumption of the $n$-th PC at time $t$ are defined as:
\begin{subequations}
\begin{align}
    d_{t,n} &= d^{crit}_{t,n} + d^{curt}_{t,n}, \\
    c_{t,n} &= c^{crit}_{t,n} + c^{curt}_{t,n}, 
\end{align}
\end{subequations}
where $d^{crit}_{t,n}$ and $c^{crit}_{t,n}$ denote the energy demand and energy consumption of the $n$-th PC at time $t$ for critical load, and $d^{curt}_{t,n}$ and $c^{curt}_{t,n}$ denote the energy demand and energy consumption of the $n$-th PC at time $t$ for curtailable load. 

The critical load demands are always completely met, whereas the curtailable load consumption follows the price elasticity of demands, i.e., 
\begin{subequations}
\begin{align}
    c^{crit}_{t,n} &= d^{crit}_{t,n}, \\
    c^{curt}_{t,n} &= d^{curt}_{t,n} \times (1 + \xi_t \times \frac{\phi_{t,n} - \pi_t}{\pi_t}), 
\end{align}
\end{subequations}
where $\xi_t < 0$ denotes the elasticity coefficient at time $t$ and $\pi_t$ represents the wholesale electricity price at time $t$.

\item {\bf Reward}:
To characterize the reward, we first define the SP's profits: $P(t) = \sum_{n=1}^N (\phi_{t,n} - \pi_t) \times c_{t,n}$. Then, we define PCs' costs $C(t)$, which is defined as the sum of the electricity costs and the dissatisfaction costs for all PCs, i.e., $C(t) = \sum_{n=1}^N (\phi_{t,n} \times c_{t,n} + \delta_{t,n})$. Here $\delta_{t,n} = \frac{\alpha_n }{2} (d^{curt}_{t,n} - c^{curt}_{t,n})^2 + \beta_n (d^{curt}_{t,n} - c^{curt}_{t,n})$ denotes the dissatisfaction cost of the $n$-th PC at time $t$, where $\alpha_n$ is the customer preference parameter \cite{yu2017incentivebased} and $\beta_n$ is a predefined constant \cite{yu2016supplydemand}. 

The reward of the dynamic pricing DR model is then defined as a weighted combination of $P(t)$ and $-C(t)$: 
\begin{equation}
    r(s_t, a_t) = \rho \times P(t) - (1-\rho) \times C(t). 
\end{equation}
Based on the reward, the total discounted return of the dynamic pricing DR model in a complete trajectory from time $t$ onward can be represented as:
\begin{equation}
    R_t  = \sum_{k=0}^{T-t} \lambda^k r(s_{t+k}, a_{t+k}),
\end{equation}
and the performance of a stochastic policy $\pi$ can be represented as 
\begin{equation}
\eta(\pi) = \mathbb{E}_{s_0,a_0,s_1 \dots} [\sum_{t=0}^{\infty} \lambda^t r(s_t,a_t)]. 
\end{equation}
\end{itemize}
As shown in \cite{kakade2002_approximatelyoptimal}, the expected return of a new policy $\pi'$ can be expressed in terms of the advantage over the old policy $\pi$: $\eta(\pi') = \eta(\pi) + \mathbb{E}_{s\sim \rho^{\pi'}, a \sim \pi'} [A^{\pi}(s,a)]$, where $A^{\pi}(s,a) = \mathbb{E}[R_t|s_t = s, a_t = a; \pi] - \mathbb{E}[R_t|s_t = s; \pi]$ represents the advantage function and $\rho^{\pi}$ represents the unnormalized discounted visitation frequencies, i.e., $\rho^{\pi}(s) = \sum_{t=0}^{\infty}\gamma^t P(s_t = s)$.

Trust Region Policy Optimization (TRPO) \cite{schulman2015trust} employs a trust region constraint to ensure that the new policy does not deviate significantly from the old policy. Furthermore, in TRPO, the policy $\pi$ is parameterized as $\pi_{\theta}$ with the parameter vector $\theta$. For notation brevity, we use $\theta$ to represent the policy $\pi_{\theta}$. Then, the new policy $\theta'$ is updated in each iteration to maximize the expected value of the advantage function:
\begin{equation}
\begin{split}
& \max_{\theta'}\ \  \mathbb{E}_{s\sim \rho^{\theta}, a \sim \theta'} [ A^{\theta} (s,a)]  \\
& \text{s.t.} \ \ \mathbb{E}_{s\sim \rho^{\theta}} [d_{KL} (\theta', \theta)] \le \delta, 
\end{split}
\label{trpo_problem}
\end{equation}
where $\delta$ is the threshold of the distance between the old policy $\theta$ and the new policy $\theta'$; $d_{KL} (\theta', \theta)$ represents the KL divergence between the old and new policies, i.e.,
$d_{KL}(\theta', \theta) = \int_{a \in \mathcal{A}} \pi_{\theta'}(a|s) \log(\frac{\pi_{\theta'}(a|s)}{\pi_{\theta}(a|s)})da$. Though parametrizing the policy may help ease computations, as indicated in \cite{tessler2019distributional}, optimizing over parametric distributions will cause local movements in the action space and converge to a sub-optimal solution.

\section{A Nonparametric Framework to Learn Distributional Policies}
\label{ODRPO_section_has_proofs}
In this section, we first develop nonparametric constrained policy optimization framework, with the KL divergence to construct trust region. Then we derive a closed-form of the policy update and propose an efficient on-policy actor-critic algorithm to address the proposed problem.

\subsection{Problem formulation}
\label{ODRPO_framework}
In our model, we release the restrictive assumption that a policy has to follow a parametric distribution class. Instead, we consider all admissible policies within the trust region, i.e, 
\begin{equation}
\mathbb{E}_{s\sim \rho^{\pi}} [d_{KL} (\pi'(\cdot|s), \pi(\cdot|s))] \le \delta.
\label{metricset}
\end{equation}
That is, the policy update is nonparametric. As long as its expected distance from the old policies is no more than a threshold level $\delta$, the policy will be considered as the candidate of the next policy update. 

Since the optimization goal in each policy update is to obtain the maximal expected value of the advantage function, the proposed model focuses on identifying the optimistic policy that falls in the set depicted in \eqref{metricset}. Therefore, our framework is shown as follows:
\begin{equation}
\begin{split}
& \max_{\pi' \in \mathcal{D}} \hspace{3mm} \mathbb{E}_{s\sim \rho^{\pi}, a \sim \pi'(\cdot|s)} [ A^{\pi} (s,a)] \\
& \text{where} \hspace{3mm} \mathcal{D} = \{\pi' | \mathbb{E}_{s\sim \rho^{\pi}} [d_{KL} (\pi'(\cdot|s), \pi(\cdot|s))] \le \delta \}.
\end{split}
\label{odrpo_problem}
\end{equation}
This framework is related to 
distributionally robust optimization (DRO)~\cite{rahimian2019distributionally}. However, we note that our constrained policy optimization is conceptually different from DRO.
Constrained policy optimization seeks the most optimistic policy that falls within a trust region, whereas DRO seeks to minimize some worst-case loss given by an adversarial distribution of unknown parameters within an ambiguity set. 

Before describing our main result, we adopt a mild and conventional assumption that generally holds true in most practical cases:
\begin{assumption}
Assume $A^{\pi} (s,a)$ is bounded, i.e., $\sup_{a \in \mathcal{A}}{|A^{\pi} (s,a)|} < \infty$, $\forall s \in \mathcal{S}$.  
\label{bounded_A}
\end{assumption}

\subsection{Policy update}
We present the optimal policy update of our method. In this formation, both the state space and the action space can be discrete or continuous. Theorems \ref{prop_dual_problem_kl} and \ref{thm_optimal_policy_kl} show the reformulation of \eqref{odrpo_problem} and closed-form of optimal policy update. The detailed proofs of the two theorems can be found in Appendix. 
\begin{restatable}{thm}{thmdualkl}
If Assumption \ref{bounded_A} holds, then the KL trust-region constrained optimization problem in \eqref{odrpo_problem} is equivalent to the following problem: 
\begin{equation}
 \min_{\beta \ge 0} \{l_0(\beta) :=\beta\delta  + \mathbb{E}_{s\sim \rho^{\pi}} \beta \log \mathbb{E}_{a \sim \pi(\cdot|s)} [e^{A^\pi (s,a)/\beta}]\}.
\label{kl_trco_dual}
\end{equation}
\label{prop_dual_problem_kl}
\end{restatable}
\begin{restatable}{thm}{thmoptpolicykl}
If Assumption \ref{bounded_A} holds and $\beta^*$ is the global optimal solution to \eqref{kl_trco_dual}, then the optimal policy solution to the KL trust-region constrained optimization problem in \eqref{odrpo_problem} is: 
\begin{equation}
    \pi'^*(a|s) = \mathbb{F}(\pi) = \frac{e^{ A^\pi (s,a)/\beta^*} \pi(a|s) }{\mathbb{E}_{a \sim \pi(\cdot|s)} [e^{ A^\pi (s,a)/\beta^*}]},  \hspace{3mm} \forall s \in \mathcal{S}, a \in \mathcal{A}.
\label{continuous_KL_optimal}
\end{equation}
\label{thm_optimal_policy_kl}
\end{restatable}
We use gradient-based global optimization algorithms \cite{wales1998_basin_hopping, zhan2006_monte_carlo_basin, leary2000_global_optimization} to find the global optimal $\beta^*$ in \eqref{kl_trco_dual}.
The gradient of the objective in \eqref{kl_trco_dual} is derived as below:
\begin{equation*}
\begin{split}
    \frac{\partial l_0(\beta)}{\partial \beta} = \delta + \mathbb{E}_{s \sim \rho^\pi}\{ \log \mathbb{E}_{a \sim \pi(\cdot|s)} [e^{A^\pi (s,a)/\beta}] \nonumber\\
    - \frac{ \mathbb{E}_{a \sim \pi(\cdot|s)} [e^{A^\pi (s,a)/\beta} \times A^\pi (s,a)]}{\beta \mathbb{E}_{a \sim \pi(\cdot|s)} [e^{A^\pi (s,a)/\beta}]}   \}. 
\end{split}
\end{equation*}
In implementation, we obtain the global optimal $\beta^*$ using the Basin Hopping algorithm \cite{wales1998_basin_hopping}, which is a two-phase global optimization method that utilizes the gradient information to speed up the search in the local phase.   

\subsection{On-policy actor-critic algorithm}
\label{drtrpo_alg_sec}
We provide a practical on-policy actor-critic algorithm as described in Algorithm \ref{odrpo_algorithm}, to address the proposed nonparametric constrained policy optimization problem. 
\begin{algorithm}
\SetAlgoLined
Input: number of iterations $K$, learning rate $\alpha$ \\
Initialize policy $\pi_0$ and value network $V_{\psi_0}$ with random parameter $\psi_0$\\
 \For{$k = 0,1,2 \dots K$}{
  Collect trajectory set $\mathcal{D}_k$ on policy $\pi_k$
  \\
  For each timestep $t$ in each trajectory, compute total returns $G_t$  and estimate advantages $\hat{A}_t^{\pi_k}$
  \\
  Update value:  $\begin{aligned} \psi_{k+1} \xleftarrow[]{} \psi_{k} - \alpha \nabla_{\psi_{k}} \sum (R_t - V_{\psi_k}(s_t))^2 \end{aligned}$
  \\
  Update policy: 
  $\pi_{k+1} \xleftarrow[]{} \mathbb{F} (\pi_k)$ via \eqref{continuous_KL_optimal} with $\hat{A}_t^{\pi_k}$ \\
 }
 \caption{On-policy actor-critic algorithm}
 \label{odrpo_algorithm}
\end{algorithm}
The trajectories sampled in the algorithm can either be complete or partial. If it is complete, $G_t$ can be obtained by using the accumulated discounted rewards, i.e., $R_t  = \sum_{k=0}^{T-t} \lambda^k r_{t+k}$. If it is partial, $G_t$ can be estimated by using the multi-step temporal difference (TD) methods \cite{de2018multi}: $\hat{R}_{t:t+n} = \sum_{k=0}^{n-1} \lambda^k r_{t+k} + \lambda^n V(s_{t+n})$. To estimate the advantage $\hat{A}_t^{\pi_k}$, we can either use the Monte Carlo approach, i.e., $\hat{A}_t^{\pi_k}=G_t - V_{\psi_k}(s_t)$ or Generalized Advantage Estimation (GAE) \cite{schulman2015high}. 

\section{Experiments}
In this section, we test the performance of our approach by comparing with multiple state-of-the-art RL approaches, for both continuous and discrete action spaces. Q-learning, A2C, TRPO, PPO, and DDPG \cite{lillicrap2015_ddpg} are among the benchmark methods we choose. The reason that we compare our method with A2C is because that it is also an on-policy actor-critic method that utilizes the advantage information. We also choose PPO as the benchmark since it is a variant of TRPO: Instead of restricting the KL divergence between the old and the
new policies, it penalizes on the KL divergence. DDPG is chosen because previous work \cite{song2022_ddpg_dr} demonstrates its effectiveness in solving DR problems. We test Q-learning only for the discrete action case and DDPG only for continuous action case, due to the nature of these algorithms: One only works for discrete actions while the other primarily supports continuous actions. The environment is implemented with OpenAI Gym \cite{brockman2016_openai_gym} and experiments are conducted using Python code, 2.7GHz quad-core intel core i7 processor and 16GB RAM hardware. 

\subsection{Experimental setup}
\begin{figure}[H]
\centering
  \begin{subfigure}{0.45\columnwidth}
    \centering
    \includegraphics[width=0.8\linewidth]{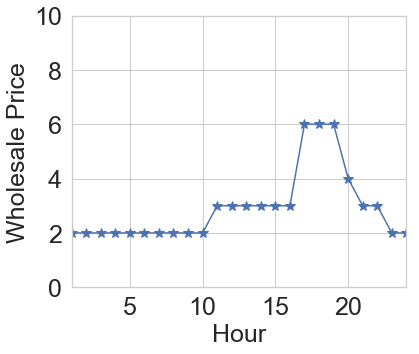}
    \caption{Wholesale price}
    \label{fig:em_wholesale_price}
  \end{subfigure}
   \begin{subfigure}{0.45\columnwidth}
    \centering
    \includegraphics[width=0.8\linewidth]{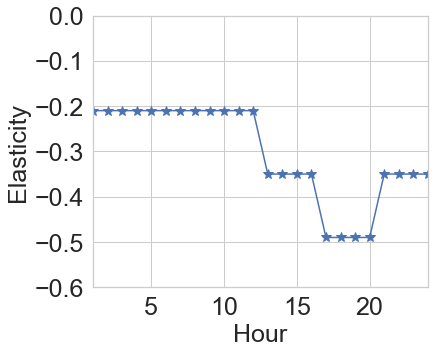}
    \caption{Elasticity}
    \label{fig:em_elasticity}
  \end{subfigure}
\caption{Wholesale price and elasticity}
\label{wholesale_elasticity}
\end{figure}

\begin{figure}[H]
\centering
  \begin{subfigure}{0.33\columnwidth}
    \centering
    \includegraphics[width=0.99\linewidth]{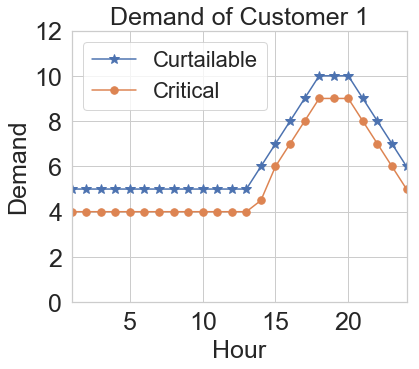}
    \caption{Customer 1} \label{fig:load_1}
  \end{subfigure}%
  \hfill
  \begin{subfigure}{0.33\columnwidth}
    \centering
    \includegraphics[width=0.99\linewidth]{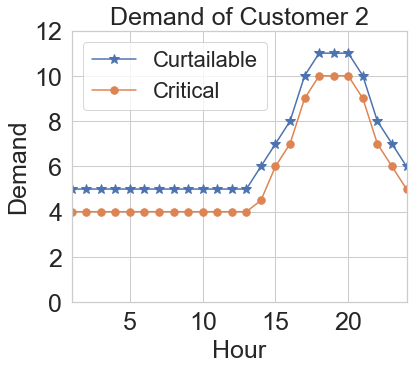}
    \caption{Customer 2} \label{fig:load_2}
  \end{subfigure}%
  \hfill
  \begin{subfigure}{0.33\columnwidth}
    \centering
    \includegraphics[width=0.99\linewidth]{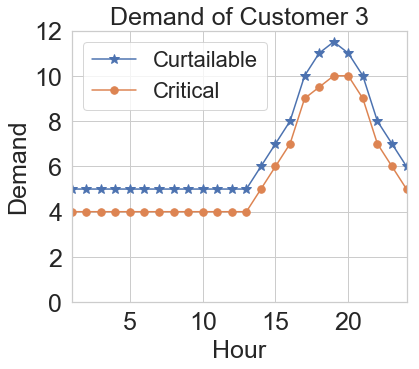}
    \caption{Customer 3} \label{fig:load_3}
  \end{subfigure}%
\caption{Customer demands within an episode}
\label{customer_demand}
\end{figure}

First, we test a simple DR case with three PCs for illustration purpose. In each episode with $24$ hours, the wholesale price $\pi_t$, elasticity coefficient $\xi_t$, critical and curtailable demands $d_{t,n}^{crit}$ and $d_{t,n}^{curt}$ are depicted in Fig. \ref{wholesale_elasticity} and \ref{customer_demand}. {Wholesale price is set to be higher, and elasticity to be lower during peak hours. }

\subsection{Performance and final policy}
In this subsection, we present the performance and final policy of our approach versus baseline algorithms on the electricity market models. For the continuous action model, any retail prices ranging from $0$ to $12$ are allowed. For the discrete action model, we adopt a discrete action space where the retail prices can only take discrete values among $0, 0.5, 1, \dots, 11.5, 12$. 

\begin{figure}[H]
\centering
  \begin{subfigure}{0.45\columnwidth}
    \centering
    \includegraphics[width=0.9\linewidth]{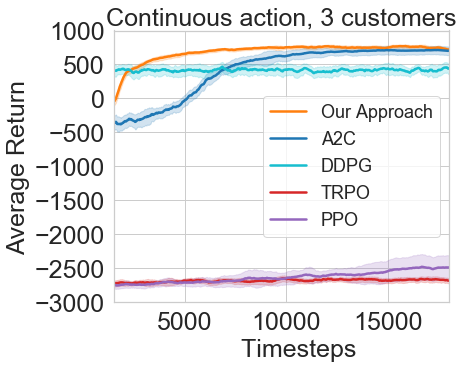}
    \caption{Continuous action model}
    \label{fig:em_performance}
  \end{subfigure}
   \begin{subfigure}{0.45\columnwidth}
    \centering
    \includegraphics[width=0.9\linewidth]{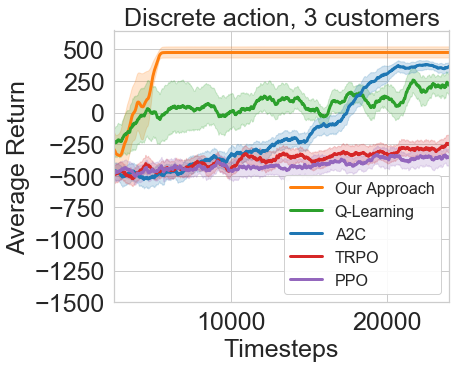}
    \caption{Discrete action model}
    \label{fig:em_discrete_performance}
  \end{subfigure}
\caption{Episode rewards during the training process, averaged across $3$ runs with a random initialization. The shaded area depicts the mean $\pm$ the standard deviation.}
\label{em_performance_continuous_and_discrete}
\end{figure}

\zred{
The performance comparisons are provided in Fig. \ref{em_performance_continuous_and_discrete}. As shown in Fig. \ref{em_performance_continuous_and_discrete}, in both continuous and discrete action model, our approach converges faster and reaches a better final performance than baseline algorithms. This suggests that the use of nonparametric policy representation indeed finds a more optimal final policy. }

\begin{figure}[H]
\centering
  \begin{subfigure}{0.33\columnwidth}
    \centering
    \includegraphics[width=0.99\linewidth]{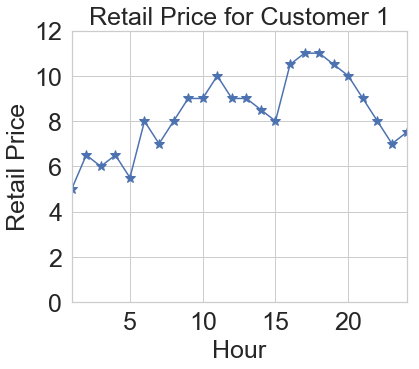}
    \caption{Customer 1} \label{fig:retail_1}
  \end{subfigure}%
  \hfill
  \begin{subfigure}{0.33\columnwidth}
    \centering
    \includegraphics[width=0.99\linewidth]{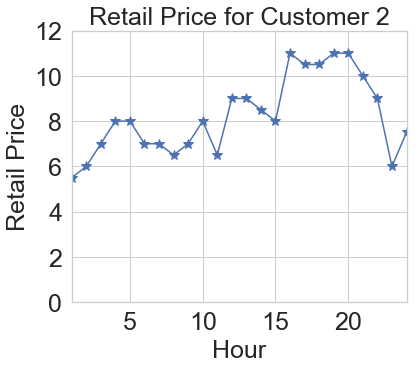}
    \caption{Customer 2} \label{fig:retail_2}
  \end{subfigure}%
  \hfill
    \begin{subfigure}{0.33\columnwidth}
    \centering
    \includegraphics[width=0.99\linewidth]{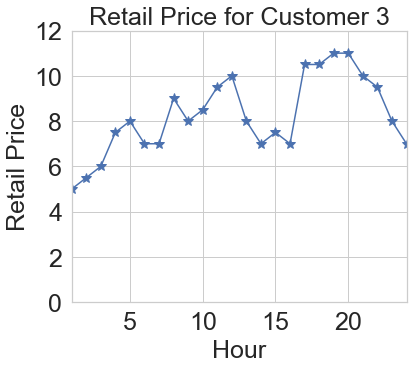}
    \caption{Customer 3} \label{fig:retail_3}
  \end{subfigure}%
 \caption{Final control actions (retail prices) for each customer within an episode, continuous action model. }
 \label{retail_price}
 \end{figure}

\begin{figure}[H]
\centering
  \begin{subfigure}{0.33\columnwidth}
    \centering
    \includegraphics[width=0.99\linewidth]{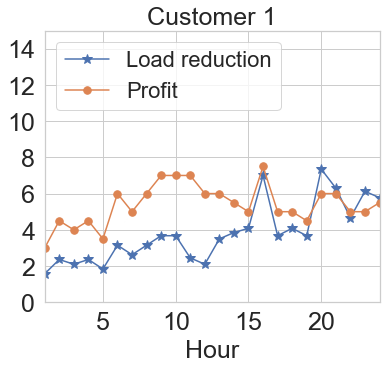}
    \caption{Customer 1} \label{fig:profit_1}
  \end{subfigure}%
  \hfill
  \begin{subfigure}{0.33\columnwidth}
    \centering
    \includegraphics[width=0.99\linewidth]{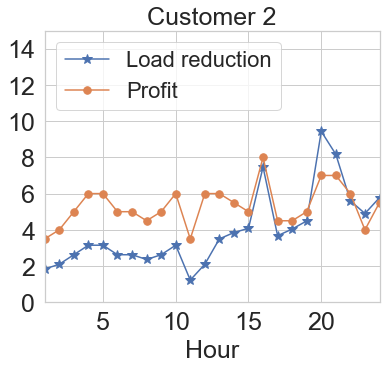}
    \caption{Customer 2} \label{fig:profit_2}
  \end{subfigure}%
  \hfill
    \begin{subfigure}{0.33\columnwidth}
    \centering
    \includegraphics[width=0.99\linewidth]{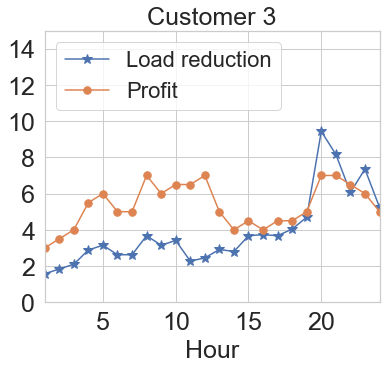}
    \caption{Customer 3} \label{fig:profit_3}
  \end{subfigure}%
 \caption{Final profit (retail minus wholesale price) and load reduction within an episode, continuous action model. }
 \label{profit}
 \end{figure}

{
We provide the final pricing strategy of our approach in Fig. \ref{retail_price}, from which we can see that the retail prices are higher during peak hours when wholesale price is high. We also analyze how this pricing strategy affects the usage in Fig. \ref{profit}, which shows that the load reduction follows a similar trend as the unit profit. Also, the load reduction is higher during peak hours, indicating that our pricing strategy effectively controls peak-hour usages. }

\subsection{Extension to large-scale action space}
\begin{figure}[H]
\centering
\includegraphics[width=0.48\linewidth]{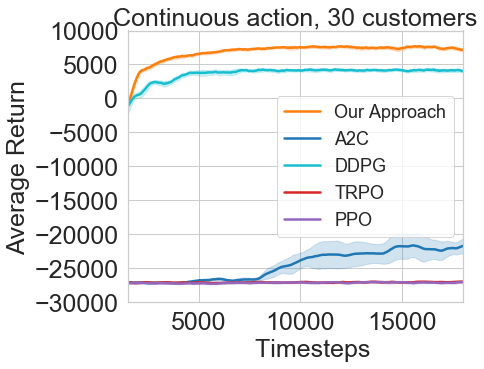}
\caption{Episode rewards during the training process, continuous action model, averaged across $3$ runs with a random initialization. The shaded area depicts the mean $\pm$ the standard deviation. }
\label{fig:em_largescale_performance}
\end{figure}

\zred{We further extend our experiments to a larger scale continuous action model with $30$ customers. We consider continuous action because it is more general and closer to practical use case. We simulate customer demands by adding small variance to Fig. \ref{customer_demand}. The performance comparison in Fig. \ref{fig:em_largescale_performance} shows similar findings as small-scale case: Our approach converges faster and has a higher performance compared to baseline algorithms. }

\section{Conclusion}
\zred{
In this paper, we present a novel nonparametric constrained policy optimization RL algorithm, which addresses the sub-optimality issue of traditional gradient based policy optimization methods such like TRPO, PPO, DDPG and A2C. Our approach improves TRPO and PPO with a better final performance, while maintaining the stable learning property. 
}

\zred{
To our knowledge, our approach is the first policy optimization method that learns nonparametric policies in DR settings. Experiments on two DR cases show the superior performance
of our proposed method compared with state-of-the-art RL algorithms.}

\bibliographystyle{unsrt}
\bibliography{pes}

%






\section{Appendix: Proofs of Theorems}
\label{appendix_proof}

\begin{proof}[Proof of Theorem \ref{prop_dual_problem_kl}]

Let $\begin{aligned} L_s(a) = \frac{\pi'(a|s)}{\pi(a|s)} \end{aligned}$. Denote $\mathbb{L}_s = \{L_s' \hspace{1mm} | \hspace{1mm} \mathbb{E}_{a \sim \pi(\cdot |s)} [L_s'(a)] = 1, \hspace{1mm} L_s' \ge 0\}$. It's easy to prove that $L_s \in \mathbb{L}_s$. By using the importance sampling and the definition of KL divergence, we have:
$\mathbb{E}_{ a \sim \pi'(\cdot|s)} [ A^{\pi} (s,a)] = \mathbb{E}_{a \sim \pi(\cdot|s)} [ A^{\pi} (s,a) L_s(a)]$, and $d_{KL} (\pi'(\cdot|s), \pi(\cdot|s)) = \mathbb{E}_{a \sim \pi'(\cdot|s)} [\log{L_s(a)}] = \mathbb{E}_{a \sim \pi (\cdot|s)} [L_s(a) \log{L_s(a)}]$.

Thus, we can reformulate \eqref{odrpo_problem} with KL divergence based trust region as: 

\begin{equation}
\begin{split}
& \max_{L_s \in \mathbb{L}_s} \hspace{3mm} \mathbb{E}_{s\sim \rho^{\pi}} \mathbb{E}_{a \sim \pi(\cdot|s)} [ A^{\pi} (s,a) L_s(a)]  \\
& s.t. \hspace{3mm} \mathbb{E}_{s\sim \rho^{\pi}} \mathbb{E}_{a \sim \pi(\cdot|s)} [L_s(a) \log L_s(a)] \le \delta. 
\end{split}
\label{continuous_KL_eq2}
\end{equation}

First, it is easy to prove that for $\forall s, a$, $\mathbb{E}_{a \sim \pi(\cdot |s)}[L_s(a)]$ and $\mathbb{E}_{s\sim \rho^{\pi}} \mathbb{E}_{a \sim \pi(\cdot|s)} [ A^{\pi} (s,a) L_s(a)]$ are linear functions of $L_s(a)$, and $\mathbb{E}_{s\sim \rho^{\pi}} \mathbb{E}_{a \sim \pi(\cdot|s)} [L_s(a) \log L_s(a)]$ is a convex function of $L_s(a)$. In addition, Slater's condition holds for \eqref{continuous_KL_eq2} since there is an interior point $L_s(a) = 1$  $\forall s,a$. Meanwhile, since $A^{\pi} (s,a)$ is bounded following from Assumption \ref{bounded_A}, the objective is bounded above. Therefore, strong duality holds for \eqref{continuous_KL_eq2}. To reformulate \eqref{continuous_KL_eq2}, we consider its Lagrangian duality function: \\
\begin{equation*}
    \begin{split}
    l_0(\beta, L_s) &=   \mathbb{E}_{s\sim \rho^{\pi}}  \mathbb{E}_{a \sim \pi(\cdot|s)} [ A^{\pi} (s,a) L_s(a)]   \\ & \hspace{10mm} - \beta \{\mathbb{E}_{s\sim \rho^{\pi}} \mathbb{E}_{a \sim \pi(\cdot|s)} [L_s(a) \log L_s(a)] - \delta \} \\
    & =   \mathbb{E}_{s\sim \rho^{\pi}} \mathbb{E}_{a \sim \pi(\cdot|s)}[ A^{\pi} (s,a) L_s(a) \\ & \hspace{10mm} - \beta L_s(a) \log L_s(a)] +  \beta\delta,
    \end{split}
\end{equation*}

where $\beta$ is the dual variable. Then, \eqref{continuous_KL_eq2} is equivalent to its dual problem as follows: 
\begin{equation}
    \min_{\beta \ge 0} \hspace{1mm} \max_{L_s \in \mathbb{L}_s} \hspace{1mm} l_0(\beta, L_s ).
    \label{continuous_KL_dual_eq1}
\end{equation}

The inner maximization problem of \eqref{continuous_KL_dual_eq1} is equivalent to: 
\begin{equation}
    \begin{split}
      \max_{L_s \ge 0} \hspace{3mm} \mathbb{E}_{s\sim \rho^{\pi}} &\mathbb{E}_{a \sim \pi(\cdot|s)} [ A^{\pi} (s,a) L_s(a) - \beta L_s(a) \log L_s(a) ] + \beta\delta 
     \\ & s.t. \hspace{3mm} \mathbb{E}_{a \sim \pi(\cdot|s)} [L_s(a)] = 1, \hspace{5mm} \forall s \in \mathcal{S}.
    \end{split}
    \label{continuous_KL_dual_eq2}
\end{equation}

By Theorem 1 of \cite{hu2012_dro_kl}, we can obtain the optimal solution $L_s^*$ and the optimal objective value of the inner maximization problem \eqref{continuous_KL_dual_eq2} respectively as follows:
\begin{equation*}
    L_s^* (a) = \frac{e^{ A^\pi (s,a)/\beta}}{\mathbb{E}_{a \sim \pi(\cdot|s)} [e^{A^\pi (s,a)/\beta}]}, 
\label{inner_L_solution}
\end{equation*}

\begin{equation*}
l_0(\beta, L_s^* ) = \beta\delta + \mathbb{E}_{s\sim \rho^{\pi}} \beta \log \mathbb{E}_{a \sim \pi(\cdot|s)} [e^{A^\pi (s,a)/\beta}]. 
\end{equation*}

Therefore, \eqref{continuous_KL_dual_eq1} can be further reformulated as: 
\begin{equation*}
     \min_{\beta \ge 0} l_0(\beta, L_s^* ) = \min_{\beta \ge 0} \{\beta\delta  + \mathbb{E}_{s\sim \rho^{\pi}} \beta \log \mathbb{E}_{a \sim \pi(\cdot|s)} [e^{A^\pi (s,a)/\beta}]\}.
\end{equation*}
\end{proof}

\begin{proof}[Proof of Theorem \ref{thm_optimal_policy_kl}]
Based on the proof of Theorem \ref{prop_dual_problem_kl}, the optimal solution $L_s^* (a)$ to \eqref{continuous_KL_dual_eq1} is: $\begin{aligned}L_s^* (a) = \frac{e^{ A^\pi (s,a)/\beta^*}}{\mathbb{E}_{a \sim \pi(\cdot|s)} [e^{A^\pi (s,a)/\beta^*}]} \end{aligned}$. Since $\pi'^*(a|s) = L_s^* (a)\pi(a|s)$, we have \eqref{continuous_KL_optimal} holds.
\end{proof}

\end{document}